\documentclass[letterpaper, 10 pt, conference]{ieeeconf}  

\IEEEoverridecommandlockouts                              

\usepackage{graphicx}
\usepackage{cite}
\usepackage{gensymb}

\title{\LARGE \bf
Soft Surfaced Vision-Based Tactile Sensing for Bipedal Robot Applications
}

\author{Jaeeun Kim$^{1}$, Junhee Lim$^{2}$, and Yu She$^{1}$
\thanks{$^{1}$Jaeeun Kim and Yu She are with the Department of Industrial Engineering, Purdue University, West Lafayette, IN 47906, USA {\tt\small \{kim2592, shey\}@purdue.edu}}%
\thanks{$^{2}$Junhee Lim is with the Department of Computer Science, Purdue University, West Lafayette, IN 47906, USA {\tt\small lim347@purdue.edu}}%
}

\begin{document}

\maketitle
\thispagestyle{empty}
\pagestyle{empty}

\begin{abstract}

Legged locomotion benefits from embodied sensing, where perception emerges from the physical interaction between body and environment. We present a soft-surfaced, vision-based tactile foot sensor that endows a bipedal robot with a skin-like deformable layer that captures contact deformations optically, turning foot–ground interactions into rich haptic signals. From a contact image stream, our method estimates contact pose (position and orientation), visualizes shear, computes center of pressure (CoP), classifies terrain, and detects geometric features of the contact patch. We validate these capabilities on a tilting platform and in visually obscured conditions, showing that foot-borne tactile feedback improves balance control and terrain awareness beyond proprioception alone. These findings suggest that integrating tactile perception into legged robot feet improves stability, adaptability, and environmental awareness, offering a promising direction toward more compliant and intelligent locomotion systems. For the supplementary video, please visit: https://youtu.be/ceJiy9q\_2Aw
 
\end{abstract}

\section{INTRODUCTION}

Legged robots are increasingly expected to walk, balance, and interact safely in complex, partially unknown environments. In these settings, maintaining stable contact with the ground is critical, especially for bipedal systems whose whole-body stability depends on a relatively small support area at the feet. Reliable estimation of contact state, not just “is the foot touching the floor,” but how it is contacting, whether it is slipping, and what it is stepping on, is central to preventing falls and enabling adaptive balance control during locomotion.

Current biped controllers typically rely on inertial measurement units (IMUs) and six-axis force–torque sensors. These traditional sensors offer high bandwidth and can be integrated into relatively simple dynamic models \cite{external_force}. Because IMUs are typically mounted on the torso or upper leg rather than at the point of contact, controllers often implicitly assume no slip or unexpected rotation at the foot during stance, which can introduce bias and drift over time in real terrain \cite{hipankle}\cite{pronto}. When mounted at the ankle or foot, force–torque sensors can detect contact events, estimate ground reaction forces, and are commonly used to compute the zero-moment point (ZMP), a key metric in dynamic gait and balance control \cite{ft_review}. However, conventional force–torque sensing provides only low-dimensional, lumped signals: it reports net force and torque, but not where within the sole contact is occurring, not the local contact geometry, and not whether the surface is about to slip. As a result, the controller is often forced to assume planar, no-slip contact, which breaks down in realistic environments.

Soft contact interfaces offer an alternative. A compliant sole can physically conform to the ground and encode local shape in its deformation, turning contact itself into an information source. Vision-based tactile sensors such as GelSight have shown that imaging soft material deformation can recover fine-grained pressure and shear, and this approach has proved robust in manipulation tasks. Extending this sensing paradigm from robot hands to robot feet is especially appealing for locomotion.

Humans already exploit this principle. Dense sensory receptors in the sole of the foot provide immediate feedback about subtle changes in pressure distribution, shear, and surface orientation, which is essential for maintaining balance and posture during standing and locomotion \cite{sasagawa2009balance}\cite{park2023balance}. Tactile feedback also enables rapid detection of unexpected environmental structure, such as the edge of a staircase or a small obstacle under the feet, even when it is not visible \cite{ushiyama2023feetthrough}. This motivates giving bipedal robots a similarly rich, localized “sense of foot feeling,” rather than requiring them to infer contact indirectly.

In this work, we embed a soft-surfaced, vision-based tactile sensor into each foot of a bipedal robot. The sensing surface is a soft silicone pad that elastically deforms on contact. An onboard camera images this deformation, producing a dense map of local pressure, shear, and contact geometry at the foot–ground interface. Unlike rigid force–torque sensing, our soft tactile sole is used not only for contact interpretation but also for real-time balance stabilization in a walking biped, treating softness itself as an information channel for control rather than just passive compliance. Our primary contributions include:

\begin{itemize}
\item A soft, foot-mounted, vision-based tactile sensor for bipedal robots. The sensor consists of a compliant silicone pad whose deformation under load is imaged internally, producing dense contact information (pressure, shear, local geometry) at the foot–ground interface.

\item A contact-state estimation pipeline. We infer depth of contact geometry, center of pressure (CoP), object pose (position and orientation) under the foot, shear distribution, and terrain class directly from tactile images.

\item Closed-loop balance control using tactile feedback. We integrate these tactile estimates into the robot’s controller and demonstrate real-time stabilization on a tilting platform.

\end{itemize}

\section{Background and Related Work}

Early efforts to equip robot feet with tactile perception focused on detecting contact events and terrain boundaries rather than reconstructing full contact state. For quadruped robots, Stone et al. \cite{stone2020walking} replaced one leg with a vision-based tactile sensing module to help the robot traverse uneven terrain. Their design could detect path edges during walking, but it did not estimate contact force magnitude or direction.

Subsequent work moved toward capturing spatially distributed contact rather than just binary contact cues. Tako et al. \cite{tako_array} embedded dense arrays of force-sensitive resistors (FSRs) in the foot to sense partial and distributed load. This provided a coarse map of the contact patch, but it did not recover detailed shape or depth information at the interface.
More recent approaches have begun to use vision-based tactile sensing to infer richer contact properties. Shi et al. \cite{shi2024foot} developed a vision-based tactile sensor designed to integrate with commercial quadruped robot feet. Their system estimated six-dimensional contact forces and identified terrain type and stiffness. However, it did not reconstruct high-resolution contact geometry, and it was evaluated on quadruped platforms rather than bipeds.

Other work has explored visuotactile sensing on reduced testbeds. Zhang et al. \cite{zhang2021tactile} used vision-based tactile input with learned models for pose estimation on a single robotic leg, demonstrating feasibility but not full-body closed-loop control. Guo et al. \cite{guo2020soft} proposed a multisensor-integrated robotic foot capable of terrain classification using multiple analog sensing modalities, though it did not incorporate visual sensing.

Vision-based tactile sensors such as GelSight use a deformable elastomer observed by an internal camera under controlled illumination to infer surface contact geometry with high spatial resolution \cite{gelorigin}\cite{visuotactile}. A typical GelSight sensor consists of a soft elastomer layer coated with a reflective surface. When the elastomer is pressed against an object, it conforms to the object’s geometry; the internal camera images that deformation, and photometric methods reconstruct local surface depth.

Unlike conventional discrete tactile elements or lumped force–torque sensing, GelSight-style visuotactile sensing can recover fine-grained 3D geometry, track lateral displacement (shear), and detect slip onsets by observing how the contact patch deforms and shears over time \cite{wang2021gelsight}\cite{9762175}. As a result, GelSight variants have been successfully used for robotic manipulation tasks such as in-hand localization, material/texture recognition, and estimation of surface properties \cite{gelorigin}. Many implementations place the sensor on fingertips or grippers and even mold the elastomer into curved, finger-like geometries to better mimic human touch \cite{10609863}\cite{9018215}.

However, despite their versatility, most visuotactile sensors have been designed for hands, not feet. Prior work generally assumes intermittent, quasistatic contacts typical of grasping, rather than continuous, load-bearing contact under body weight during locomotion. To our knowledge, no prior work has demonstrated a GelSight-style sensor adapted as a load-bearing, compliant foot sole that provides dense contact geometry and shear information while simultaneously contributing to real-time balance control in a legged robot. This motivates extending visuotactile sensing from manipulation to locomotion: in our design, the soft silicone pad on the sole acts both as a compliant physical interface (spreading load, conforming to small terrain features) and as a high-resolution sensing surface whose deformation is directly interpretable for feedback control.

\section{Hardware Design}
\subsection{Biped Robot Platform}

To implement and evaluate our tactile feedback control system, we utilized the Hash Humanoid Robot V3, an open-source bipedal platform developed by Hash Robotics \cite{hashv3}. The robot is equipped with 18 MG995 servo motors, of which 12 articulate the legs. It weighs 1.6 kg and stands 40 cm tall when fully upright, including the custom foot-mounted GelSight sensor.

The platform is capable of performing basic motions such as standing, walking, and arm waving, but lacks onboard proprioceptive or exteroceptive sensing. The ankle joints provide a ±30° rotation range, enabling moderate postural adjustments. Our sensor was mechanically designed to match the robot’s foot profile, resulting in a compact, block-like form factor that integrates seamlessly with the platform.

\subsection{Sensor} 
A custom-designed GelSight sensor was mounted to the right foot. For the elastomer pad, a mixture of 10 parts Silicones, Inc.'s P-565 silicone base and 1 part P-565 activator from the same manufacturer was degassed in a vacuum chamber for 15 minutes, then poured into a mold to form a pad measuring 58 mm \texttimes{} 90 mm \texttimes{} 3 mm. A 5 mm-thick acrylic plate was placed over the mold while the silicone cured at room temperature for 24 hours.

After curing, the surface of the elastomer was coated with Paint-On\textsuperscript{\textregistered} gray silicone paint, prepared by mixing 1 part gray silicone ink, 10 parts silicone base, and 3 parts adhesive. To improve durability and wear resistance, a layer of 3M Tegaderm\texttrademark{} medical tape was applied over the painted surface.

LED strips in four colors, red, green, blue, and white, were attached to the inner sides of the acrylic plate. Each red and green strip contained 29 5050-size LEDs, while the blue and white strips contained 18 each. The LED strips were powered by PWM outputs from an Arduino microcontroller supplying 3.3V. Three layers of optical diffusers were added above the LEDs, topped with a gray filter to ensure even illumination across the sensor surface.

\begin{figure}[thpb]
      \centering
      \includegraphics[scale=0.25]{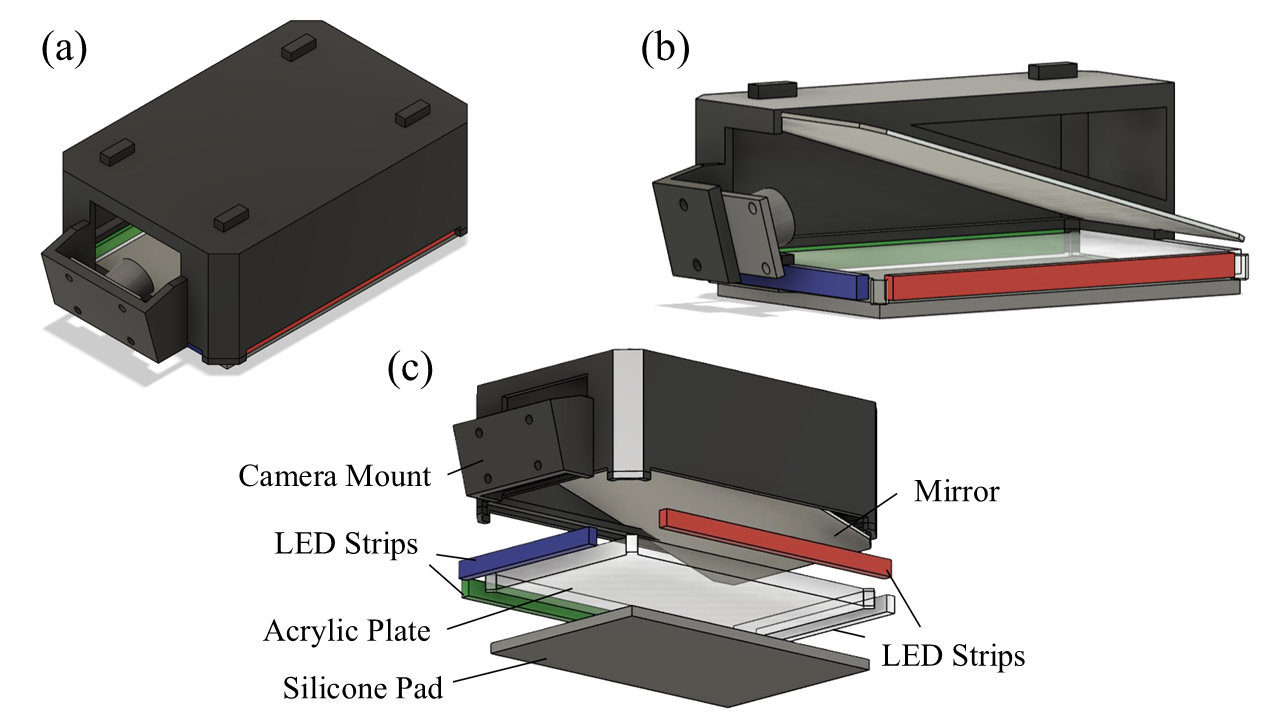}
      \caption{Foot sensor design. (a) Isometric view of the sensor. (b) Side view with partially removed housing. (c) Exploded view with labeled parts.}
      \label{sensorhardware}
   \end{figure}

The sensor housing (Figure \ref{sensorhardware}) measures 10 cm \texttimes{} 7 cm \texttimes{} 4 cm including the silicone pad, with a protruding camera fixture 1.5 cm long. A mirror was attached at 15\degree{} to minimize the sensor height. The images reflected in the mirror were read with a Raspberry Pi camera with a 160\degree{} field of view. 

\section{Sensor Perception Tasks} \label{sec4}
The image data, 640 \texttimes{} 480 pixels at 30 fps, were streamed over Wi-Fi (HTTP) from the foot-mounted sensor. Each frame was cropped to a 114 \texttimes{} 143 region of interest (ROI) and dispatched to task-specific pipelines: depth perception, shear force detection, terrain classification, object position and orientation,  CoP estimation. 

\subsection{Depth Perception}
\label{depthp}
The sensor uses photometric stereo to highlight elastomer deformations under multi-color, uniform illumination \cite{gelorigin}. These visual cues enable inference of the 3D contact geometry.

We adopt a data-driven reconstruction: we built a calibration set by manually indenting the sensor with a 4 mm sphere and annotated the indentation center and diameter. A neural network then predicts per-pixel surface gradients. Specifically, a three-layer multilayer perceptron (MLP) was employed to predict gradient values ($g_x$, $g_y$) based on pixel-wise input features. Each input consisted of the RGB color values and (\textit{x}, \textit{y}) pixel coordinates, yielding an input vector of size 5. The MLP consisted of three hidden layers with 64 units each, using ReLU activation and a dropout rate of 0.1. From 5,000 total images, 80\% were used for training and 20\% for testing. To improve robustness, 566 images were removed due to noise, primarily caused by inconsistent lighting, and artificial empty points were introduced at a rate of 3\%.

The network was trained over 30 epochs with a batch size of 64 and a learning rate of 0.001. The final model achieved a mean squared error (MSE) of 0.00285. Gradient outputs were then passed to a Poisson solver to reconstruct the depth map, enabling the sensor to recover detailed geometric features of the contact surface \cite{wang2021gelsight}.

\begin{figure}[h]
\includegraphics[width=\linewidth, height=8cm]{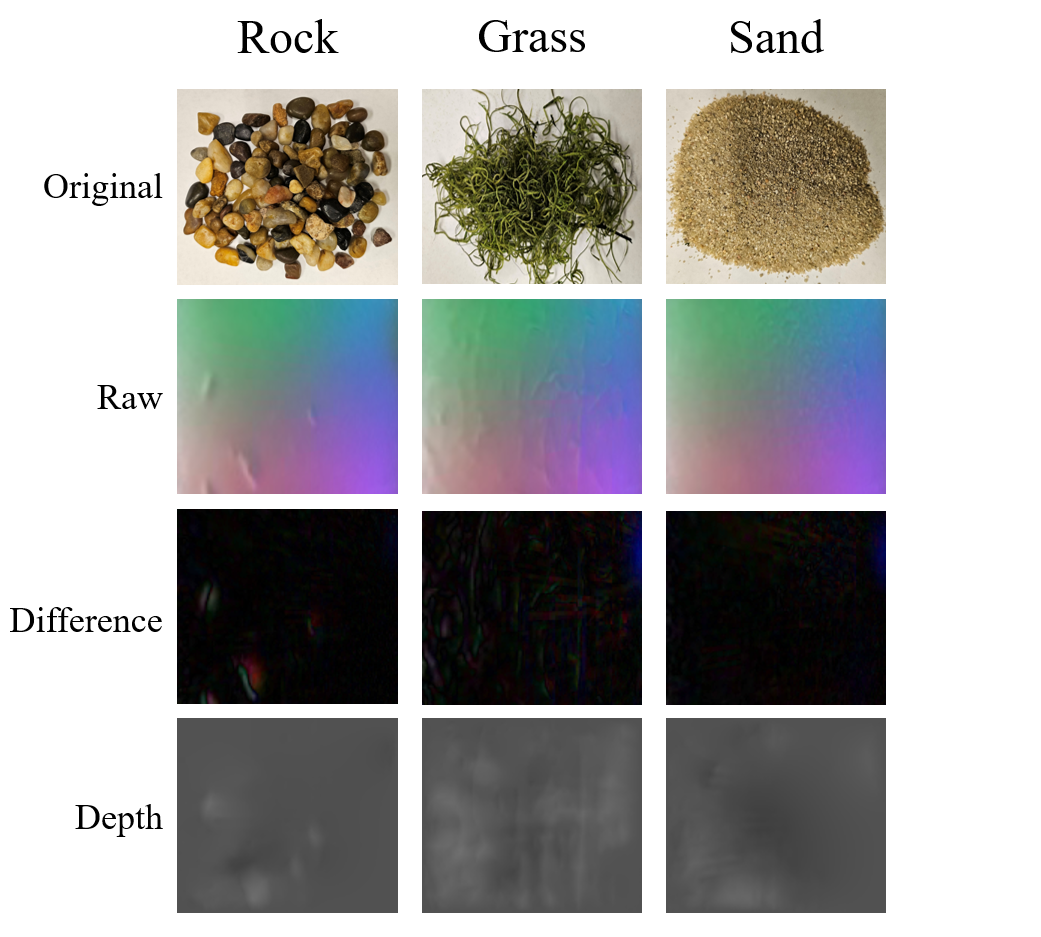} 
\centering
\caption{Depth reconstruction from the sensor using a trained model.}
\label{depthpipe}
\end{figure}

Figure \ref{depthpipe} shows the material, raw image, reference-subtracted difference, and reconstructed depth, highlighting distinct contact-induced patterns. From top to bottom, it shows the original material, the raw image captured by the sensor, the difference image computed by subtracting the reference frame, and the resulting depth image. The depth map reveals distinct geometric patterns caused by the contact deformation of each material, demonstrating the sensor's ability to reconstruct 3D surface geometry based on visual cues.

\subsection{Shear Force Detection}
The sensor was also capable of detecting shear forces by observing grid marker displacement across the surface of a modified elastomer pad. To enable this, the standard silicone surface was replaced with a version containing a 9 \texttimes{} 14 grid of printed markers and the model was trained using synthetically generated images containing a 9 \texttimes{} 14 marker grid on them, adopting the work of Wang et al.\cite{wang2021gelsight}.

\begin{figure}[h]

\includegraphics[width=\linewidth, height=4.5cm]{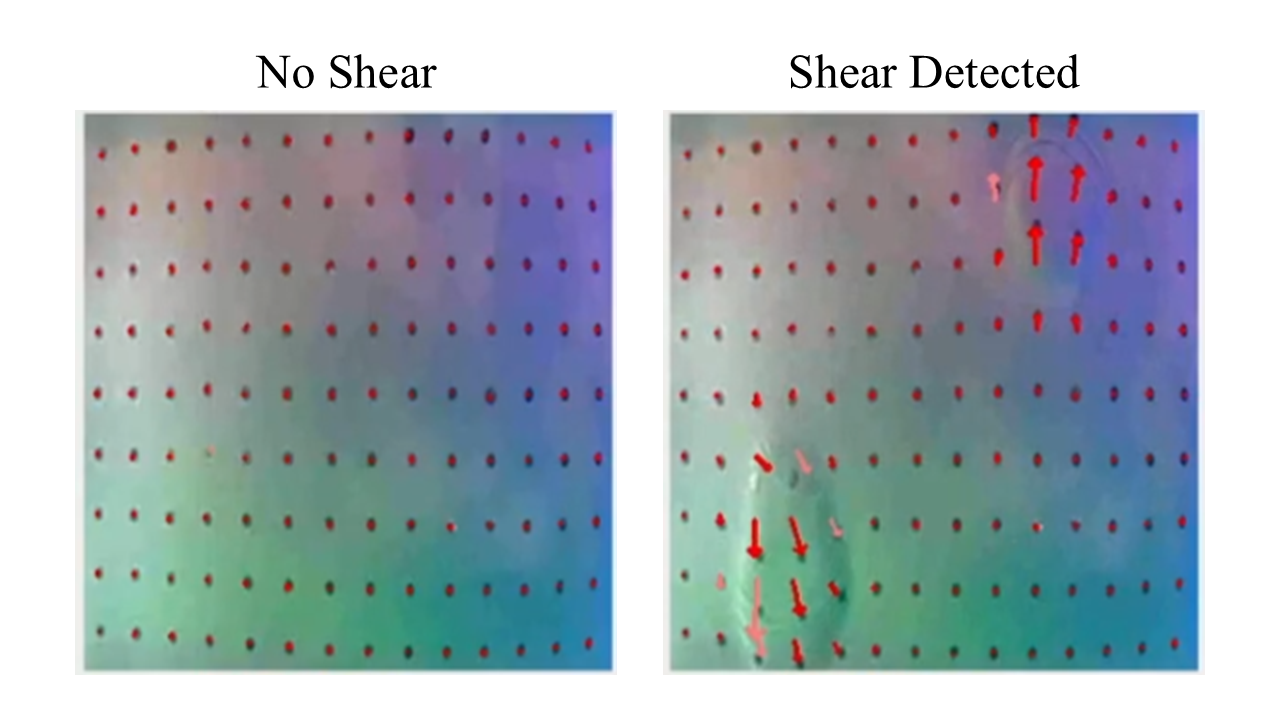} 
\caption{Detecting shear force with dotted grid pad. Shear force is represented with pink and red arrows.}
\label{shear}
\end{figure}

As the sensor experienced lateral forces, the displacement of each marker was tracked in real time. The resulting shear vectors were visualized as arrows overlaid on the sensor image, where arrow length represented the magnitude of the shear force, and the arrowhead indicated the direction. Red arrows correspond to successfully detected markers, while pink arrows indicate interpolated vectors in regions where marker detection failed but motion was inferred from previous frames. Figure \ref{shear} illustrates both a neutral state (no shear force) and an active shear detection scenario. This visual feedback demonstrates the sensor’s ability to detect shear forces, which are essential for identifying slip or tangential loads during contact. 

\subsection{Terrain Classification}
\label{ter_class}
The image classification task was implemented using a transfer learning approach with ResNet-50 as the backbone. The objective was to distinguish among four terrain categories in the collected dataset. The network was initialized with ImageNet pretrained weights and modified by replacing the final fully connected layer with a four-class output.

The model architecture followed the standard ResNet-50 design of residual blocks with batch normalization and ReLU activations. The final global average pooling layer was preserved to reduce overfitting while maintaining spatial context. The classification head was adjusted to match the target classes, and label smoothing with a factor of 0.1 was applied to improve calibration and generalization.

Training was performed using the AdamW optimizer with a base learning rate of 0.001 and a weight decay of 0.05. A OneCycle learning rate scheduler dynamically adjusted the learning rate over the course of training. Regularization included MixUp and CutMix augmentations, each applied with 70\% probability per batch. Early stopping was enforced when validation loss failed to improve for six consecutive epochs.

The dataset was divided into training and validation sets, following the PyTorch ImageFolder convention. There were 1042 blank images, 956 rock images, 947 spike images, and 995 tile images for the training set. The test images were collected by removing the sensor from the foot and manually pressing on the terrain samples. For the validation set, there were 100 images each. Images were resized to 224 \texttimes{} 224 pixels and normalized using ImageNet statistics. Training was conducted for a maximum of 40 epochs with a batch size of 64. Both the model with the lowest validation loss and the final model after training were checkpointed for reproducibility.

The model with lowest validation loss had a training loss of 0.5734 and a training accuracy of 88.53\%, while having a validation loss of 0.3754 and a validation accuracy of 98.75\%. The confusion matrix on the test set collected while the robot was walking in an obscured field can be found in Section \ref{terr_ex}.

\subsection{Object Position and Orientation}
\begin{figure}[h]
\includegraphics[width=\linewidth, height=4cm]{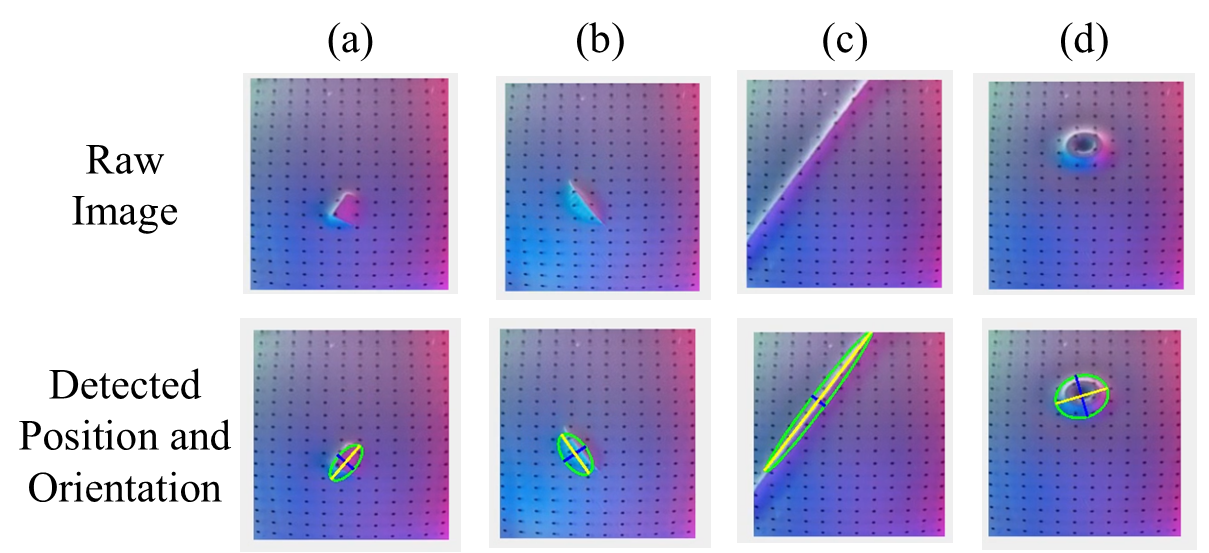}
\centering
\caption{Detecting position and orientation of the contact region of different geometries. Raw images in the upper row, detected images in the lower row. (a) Corner of a cube. (b) Circular edge of a cylinder. (c) Edge of a prism. (d) Screw head.}
\label{orientation}
\end{figure}

The sensor also enables estimation of the orientation and position of objects in contact with its surface. Figure \ref{orientation} demonstrates the sensor’s capability to estimate both the position and orientation of an object in contact. The upper row shows the raw images cropped from the stream, while the lower row illustrates the corresponding detected positions and orientations of the pressed object. The sensor identifies the edge and fits an ellipse to the contact region, with the major axis aligned along the detected edge. By computing frame differences using OpenCV, the system isolates moving contact regions and analyzes their geometric features such as contours, bounding boxes, and moments.

This approach allows the system to determine both the spatial location and angular alignment of the object in contact, which is particularly useful in tasks involving object tracking, terrain mapping, or contact shape recognition. In the presented example, the edge of the cube was accurately localized and its orientation estimated in real time based on the shape of the contact region.

\subsection{CoP Estimation}
\label{cop}
The sensor’s images were processed to estimate the CoP when contacting a textured surface. The platform used for testing was covered with a 3D-printed plastic plate featuring a grid of small cylinders, mimicking a textured floor.

\begin{figure}[h]
\includegraphics[width=\linewidth, height=3cm]{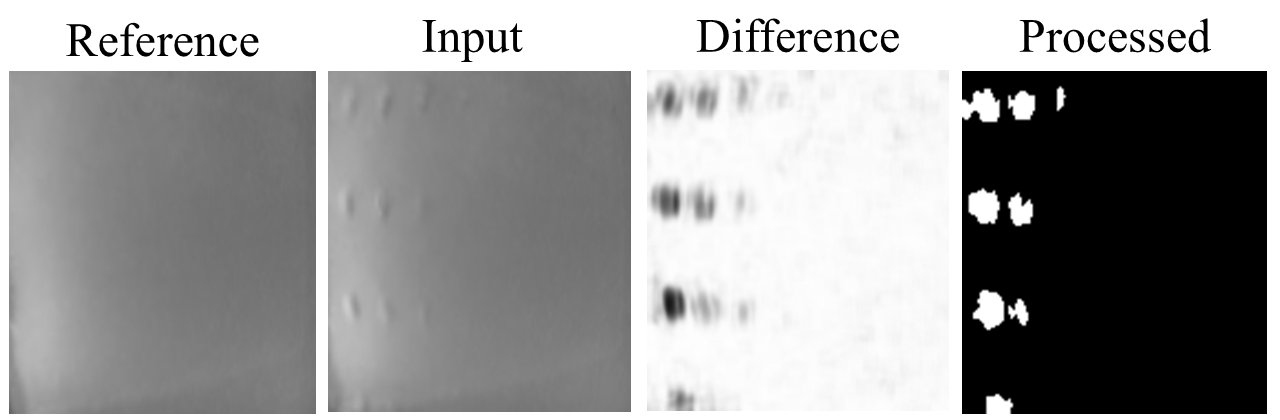}
\centering
\caption{Processed image with OpenCV function. The image processing steps flow from left to right to detect where the pressure is being applied.}
\label{cv_pipe}
\end{figure}

Figure \ref{cv_pipe} illustrates the image processing steps used to estimate the CoP from tactile sensor data. First, a reference image was captured in the absence of contact and converted to grayscale. During operation, each incoming frame was also converted to grayscale and compared to the reference image to compute a pixel-wise difference. This difference image was then thresholded using an inverted binary filter (threshold values: 200 to 255) to remove noise and isolate the deformation patterns. As pressure was applied to the sensor, circular blobs caused by the texture became visible in the image. These were identified using OpenCV’s contour detection method.

The CoP was computed by averaging the center coordinates of the detected contours. Specifically, the x-coordinate of the CoP was estimated as a visual proxy for CoP using the weighted mean:

\begin{equation}
    CoP_{x} = \frac{\sum_{i=1}^{N} a_{i}p_{i}}{\sum_{i=1}^{N} a_{i}}
\end{equation}

where $a_{i}$ is a binary weight (1 if pressure is detected at pixel \textit{i}, 0 otherwise), and $p_{i}$ is the x-position of that pixel.

\begin{figure}[h]
\centering
\includegraphics[width=7cm, height=8cm]{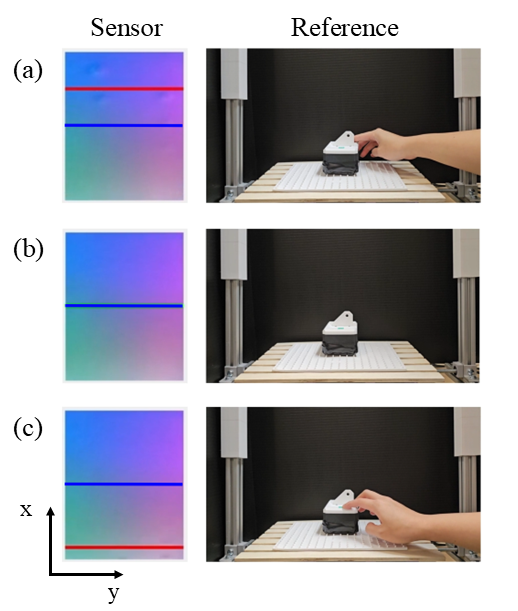} 
\label{fig:front_cop}
\caption{Estimated CoP. Red and green horizontal lines represent the \textit{x}-mean, while blue lines represent the geometric center of the sensor. The line turns red when it is outside of the safe range, and maintains green when within the safe range. (a) Pressure applied to the back of the sensor. (b) No pressure applied to the sensor. The green line indicating that the CoP is in safe area is coinciding with the blue line. (c) Pressure applied to the front of the sensor.}
\label{copestnew}
\end{figure}

Figure \ref{copestnew} displays the sensor outputs, where the blue horizontal line represents the geometric center of the sensor, and the red and green lines indicate the CoP. We used red to highlight cases where the CoP lies outside the safe range, defined experimentally as beyond 25\% of the distance from the geometric center. Figure \ref{copestnew} (b) shows a neutral contact state where the CoP coincides with the center. In Figures \ref{copestnew} (a) and (c), the sensor was pressed at the back and front respectively, resulting in a visible shift of the CoP away from the center. For CoP within the safe range, we used the color green, which is not visible in the figure but is a feature utilized in later experiments.

\section{Experiments} \label{sec5}
\subsection{Active Bipedal Balancing}
\label{experiment}
To evaluate the robot’s ability to maintain balance using tactile feedback, we conducted an experiment on a motorized tilting platform covered with a textured surface. The control objective was to minimize the error between the estimated CoP and the geometric center of the sensor, thereby stabilizing the robot's posture.

\begin{figure}[h]
\centering
\includegraphics[width=7cm, height=11cm]{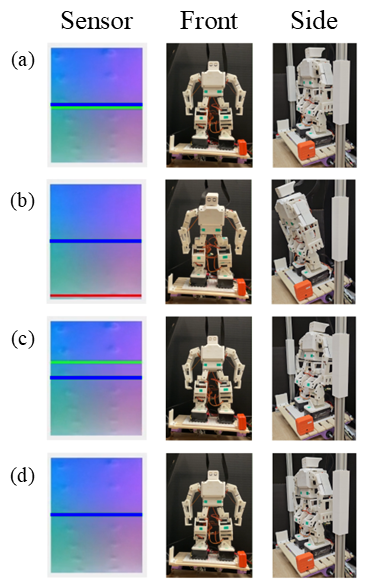} 
\caption{Robot balancing based on the estimated CoP at 0.3 rad/s speed to 15\degree{} tilted forward. The line turns red when it is outside of the safe range, and maintains green when within the safe range. Initial position depicted in (a) \textit{t} = 0 s. At (b) \textit{t} = 3 s, the robot experiences disturbance, and at (c) \textit{t} = 5 s, the robot reacts to stabilize. Robot adjusted at (d) \textit{t} = 7 s.}
\label{real_demo}
\end{figure}

\begin{figure*}[h]

\includegraphics[width=\linewidth, height=7.5cm]{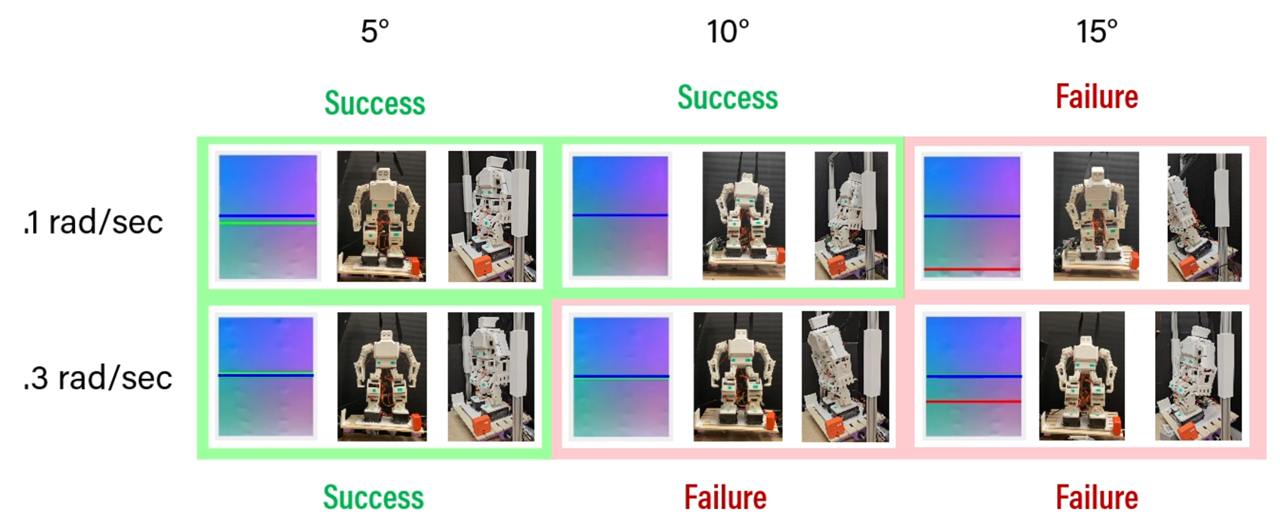} 
\caption{Summary of the experiment done on an declined (front leaning) plane. The plane was tilted with varying angles of  5°, 10°, and 15°, with different speed of 0.1 rad/s and 0.3 rad/s. The result shows that the robot was able to maintain balance with slower speed and less steep angles. For the case of 0.3 rad/s with 10°, the robot maintained its balance by harness rather than using sensor feedback, thus indicating failure.}
\label{result1}
\end{figure*}
\begin{figure*}[h]
\includegraphics[width=\linewidth, height=7.5cm]{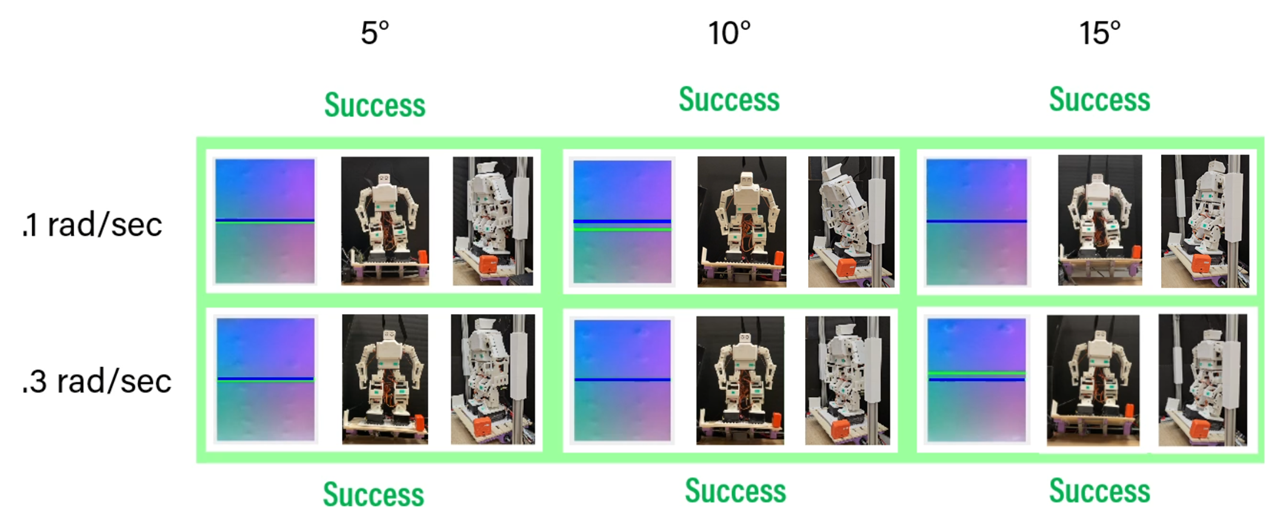} 
\caption{Summary of the experiment done on an inclined (back leaning) plane. The plane was tilted with varying angles of  -5°, -10°, and -15°, with different speed of 0.1 rad/s and 0.3 rad/s. The result shows that the robot was able to maintain balance with all cases.}
\label{result2}

\label{resutlt_two}
\end{figure*}

Sensor images were streamed and processed in real time using the OpenCV-based functions described in Section \ref{cop}. The CoP was estimated from each frame and compared to a predefined setpoint, the center of the sensor. The error \textit{e} was calculated as the difference between the estimated CoP and this setpoint.

The tactile sensor captures images at 30 frames per second (fps), which are streamed via Wi-Fi to the host PC. Each frame undergoes image preprocessing and CoP estimation in approximately 25 to 35 ms, resulting in an effective perception frequency of ~28 to 30 Hz. The control loop runs at the same rate, that is, the system computes a new ankle angle command after every processed frame. Communication with the Arduino Mega 2560 introduces minimal latency (~5 to 10 ms). Thus, the total sensing-to-actuation cycle operates at approximately 25 to 28 Hz with an estimated end-to-end delay of less than 50 ms, which is sufficient to maintain a quasi-static balance under moderate disturbances.

A proportional–integral–derivative (PID) controller was implemented to generate an ankle joint correction angle, $\theta_{off}$, based on the CoP error:

\begin{equation}
    \theta_{off} = K_p * e + K_i * e_i + K_d * e_d
\end{equation}

where the gains $K_p = 0.35$, $K_i = 0.004$, and $K_d = 0.01$ were tuned experimentally. Only the ankle joints were actuated during the experiment; the remaining joint angles were fixed to maintain an upright posture. When the error exceeded a certain threshold, the control signal was sent via serial communication to an Arduino Mega 2560, which actuated the corresponding servo motors using PWM.

The robot was tested under several platform tilt conditions, 5°, 10°, and 15°, in both forward (declining) and backward (inclining) directions, at two angular velocities: 0.1 rad/s and 0.3 rad/s. Two scenarios were compared: with tactile feedback (active PID control) and without tactile feedback (fixed ankle joints).

Figure \ref{real_demo} provides a time-series visualization of the robot’s balancing behavior over 7 seconds. At \textit{t} = 0 s, the CoP is aligned with the center. As the platform tilts forward (\textit{t} = 3 s), the CoP shifts accordingly to the front, but the robot compensates by adjusting its ankle posture (\textit{t} = 5 s), eventually restoring alignment (\textit{t} = 7 s). This real-time response highlights the ability of the PID-controlled system to dynamically stabilize the robot based on tactile input.

To quantify performance, the experiment was repeated at tilt angles of 5°, 10°, and 15°, in both forward (declined) and backward (inclined) directions, at angular velocities of 0.1 rad/s and 0.3 rad/s. The robot was considered successful if it maintained the CoP within a “safe zone” defined as ±25\% of the foot width from the center, without relying on a safety harness.

Figures \ref{result1} and \ref{resutlt_two} summarize the success rates of the robot’s balancing task under various tilt angles and directions, with and without tactile feedback. Figure \ref{result1} presents performance on a declined platform (front leaning), and Figure \ref{result2} shows results on an inclined platform (back leaning). For the declining cases, the robot was able to balance itself in 5\degree{} and 10\degree{} when rotated at 0.1 rad/s. With a faster platform movement of 0.3 rad/s, the robot was only able to balance itself at 5\degree. In both cases, successful balance is achieved at low-to-moderate tilt angles (5°, 10°) when tactile feedback is integrated. In contrast, the robot consistently fails to maintain safe CoP alignment when tactile sensing is not used. The asymmetry in performance between inclining and declining cases may be attributed to the robot’s forward weight distribution. On the other hand, while not shown in Figure \ref{resutlt_two}, all cases without tactile feedback failed. Without tactile feedback in the control loop, the robot's ankles were in fixed position along with other joints, maintaining a rigid, upright position. In those cases, even when the robot appears to be in balance without falling, the sensor showed that the robot failed to keep the CoP in the safe area.

\subsection{Terrain Classification}
\label{terr_ex}

To evaluate the sensor’s ability to classify terrains under obscured vision, we conducted an experiment in which the robot walked across a small field covered with a 0.2 mm thick white cotton fabric. Beneath the fabric are three terrain types: plastic tiles (12 \texttimes{} 12 mm, ~25 mm spacing), rocks (similar size, irregular spacing 20 to 40 mm), and 3D-printed spikes (2.5 mm diameter \texttimes{} 2.5 mm height, 12.5 mm grid).

\begin{figure}[h]
\centering
\includegraphics[width=8.5cm, height=9.2cm]{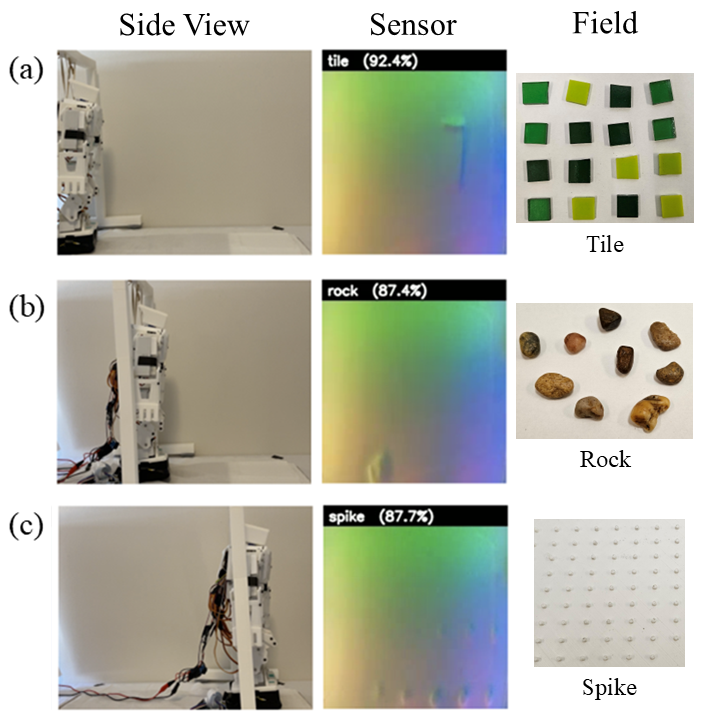} 
\caption{Terrain classification tested with the robot walking on a field covered by a white fabric. The texts on the top of the sensor images represent the class with highest confidence and the percentage next to it denotes the confidence. Objects in the third column represent the materials on the field. (a) At \textit{t} = 0 s, tile detected. (b) At \textit{t} = 30 s, rock detected. (c) At \textit{t} = 75 s, spike detected.}
\label{walking}
\end{figure}

The tactile sensor was mounted on the robot’s right foot and collected images continuously as the robot, without the prior knowledge of the field's structure, walked across the field. Figure \ref{walking} illustrates the experiment process in different time steps. As the robot walks through the path, the sensor could detect and identify impressions caused by different objects underneath the fabric. 

\begin{figure}[h]
\includegraphics[width=7cm, height=6cm]{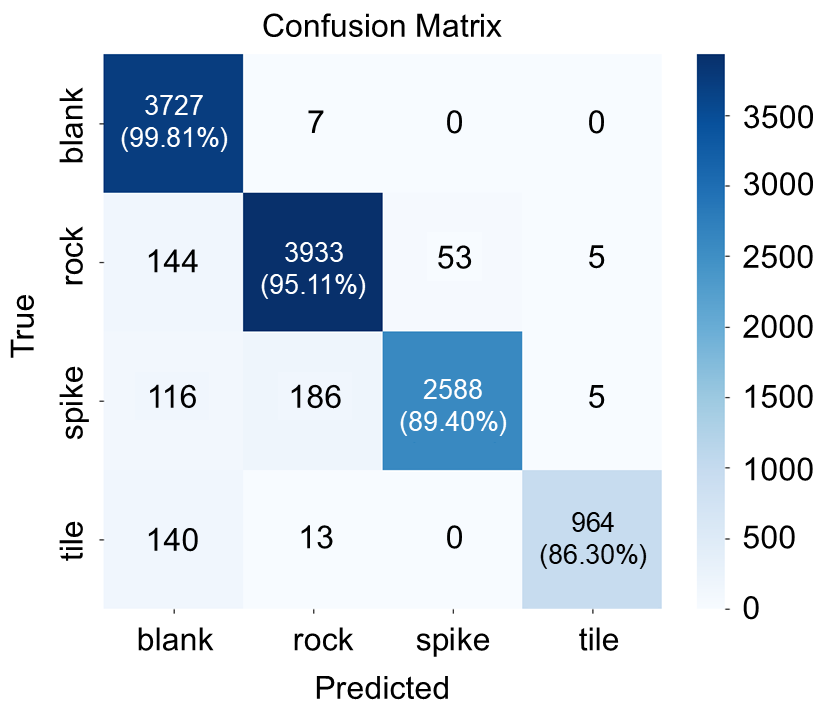} 
\centering

\caption{Confusion matrix for terrain classification using ResNet-50. Evaluated on the test set (\textit{N} = 11881); values are row-normalized percentages (left: true labels, bottom: predictions).}
\label{confusion_exp}
\end{figure}
The collected images were manually annotated as one of the terrain classes (tile, rock, or spike) or as blank when no terrain contact was present. These images were then evaluated using the same model trained in Section \ref{ter_class}. The resulting confusion matrix is presented in Figure \ref{confusion_exp}, showing that the model achieved over 85\% classification accuracy for all terrain categories.

These results demonstrate that the proposed tactile sensor can reliably classify terrain features even when visual cues are obscured by a thin fabric layer. The ability to distinguish among different terrain types under such conditions highlights the robustness of the sensor to environmental occlusion and surface variation. This finding suggests that the sensor can be applied in more realistic scenarios where direct vision is limited or compromised, such as a field covered with snow, strengthening its utility for bipedal exploration.

\section{Limitations} \label{sec6}
While our approach demonstrates the feasibility of using vision-based tactile sensing for biped stabilization, several limitations remain. First, the hardware platform used in our experiments, Hash Humanoid V3, has limited torque and actuation precision, which constrains the stability performance under more aggressive perturbations. Future work will require higher-fidelity platforms to evaluate the method under more realistic and dynamic conditions.

Second, although our terrain classification model showed high accuracy on the collected dataset, the range of terrain types was relatively limited and curated. Because of the robot's light weight, it was more difficult to detect soft materials or terrains as not enough pressure was applied when standing on the said conditions. Performance under more diverse, real-world surface conditions (e.g., wet, deformable, or cluttered terrain) remains to be evaluated.

Third, the shear vectors reported in this work are inferred from surface-marker displacements and therefore indicate relative tangential loading only. We did not perform absolute force calibration, so magnitudes are not expressed in physical units (N), and cross-sensitivities with normal pressure may remain.

Lastly, the current control policy relies on a hand-tuned PID controller. While sufficient for our demonstration, this may not scale well to more complex scenarios. Integrating tactile feedback into learning-based or model-predictive control frameworks may provide more adaptive and generalizable performance in the future.

\section{Conclusion and Future Work}\label{sec7}

In this paper, we presented a vision-based tactile sensing system for bipedal robots, integrated into a custom foot-mounted sensor. The sensor enables rich perception of contact geometry and dynamics, allowing the robot to estimate depth, CoP, shear forces, and object orientation, as well as to classify terrain types through a trained neural network.

We demonstrated that this tactile feedback can be effectively used in a closed-loop control system to stabilize a bipedal robot on a tilting platform. The robot was able to actively adjust its posture by realigning its CoP using real-time sensor input, significantly outperforming open-loop control without tactile feedback.

For future work, we plan to extend our method to dynamic walking scenarios, where continuous tactile feedback will be used to assist with step planning and gait adaptation. These capabilities may also support learning-based locomotion strategies by providing dense, high-dimensional input to reinforcement or imitation learning models.

Lastly, given the modular nature of our sensor, we intend to explore additional perception tasks, such as slip detection or surface compliance estimation, without modifying the hardware. We believe this direction can contribute to more adaptable, robust bipedal locomotion in unstructured environments.

\section{Acknowledgments}
We thank Jeonghoo Yoo and Vincent Wang for their contributions to developing the image processing pipeline and training machine learning models. 

\addtolength{\textheight}{-12cm}   





\bibliographystyle{IEEEtran}
\bibliography{main}

\end{document}